\documentclass[sigconf, nonacm, review=false, anonymous=false]{acmart}
\usepackage[utf8]{inputenc}
\usepackage{natbib}
\usepackage{graphicx}
\usepackage{todonotes}  
\usepackage{amsfonts}
\usepackage{algorithm,algorithmic}
\usepackage{enumitem}
\usepackage{adjustbox}
\usepackage{url}
\usepackage{subfig}
\usepackage{arydshln} 
\usepackage{pifont} 
\newcommand{\xmark}{\ding{55}}%

\newcommand{\Comp}{\mathbb C}
\newcommand{\Hy}{\mathbb{H}}
\newcommand{\Eu}{\mathbb{R}}
\newcommand{\EntitySub}{\mathcal{E}}
\newcommand{\RelSub}{\mathcal{R}}
\newcommand{\TripSub}{\mathcal{T}}
\newcommand{\Head}{\operatorname{\textbf{h}}}
\newcommand{\Rel}{\operatorname{\textbf{r}}}
\newcommand{\Tail}{\operatorname{\textbf{t}}}

\newcommand{\RotRef}{\textsc{RotRef}}
\newcommand{\RotRefHy}{\textsc{RotRefHy}}
\newcommand{\RotRefEu}{\textsc{RotRefEu}}
\newcommand{\MuR}{\textsc{MuR}}
\newcommand{\MuRP}{\textsc{MuRP}}
\newcommand{\MuRE}{\textsc{MuRE}}
\newcommand{\RotatE}{\textsc{RotatE}}
\newcommand{\TransE}{\textsc{TransE}}
\newcommand{\TransH}{\textsc{TransH}}
\newcommand{\DistMul}{\textsc{DistMul}}
\newcommand{\CML}{\textsc{CML++}}
\newcommand{\NeuMF}{\textsc{NeuMF}}
\newcommand{\BPR}{\textsc{BPR}}
\newcommand{\HyperML}{\textsc{HyperML}}
\newcommand{\Narre}{\textsc{Narre}}

\newlength\myindent
\copyrightyear{2021}
\acmYear{2021}
\acmDOI{10.1145/XXXXXX.XXXXXXX}
\setcopyright{none}

\acmConference[SIGIR '21]{Woodstock '18: ACM Symposium on Neural
Gaze Detection}{July 11--15, 2021}{Virtual Event, Canada}
\acmBooktitle{Proceedings of the 44th International ACM SIGIR Conference on Research and Development in Information Retrieval (SIGIR ’21), July 11--15, 2021, Virtual Event, Canada}
\acmPrice{15.00}
\acmISBN{978-1-4503-XXXX-X/18/06}

\begin{document}

\title{Augmenting the User-Item Graph with Textual Similarity Models}


\author{Federico L\'opez}
\authornote{Work partially done during an internship at Google.}
\affiliation{%
  \institution{Heidelberg Institute for Theoretical Studies}
  \city{Heidelberg}
  \country{Germany}}
  \email{federico.lopez@h-its.org}

\author{Martin Scholz}
\affiliation{%
  \institution{Google}
  \city{Mountain View}
  \country{USA}}
  \email{mscholz@google.com}

\author{Jessica Yung}
\affiliation{%
  \institution{Google}
  \city{Z\"urich}
  \country{Switzerland}}
  \email{jessicayung@google.com}
  
\author{Marie Pellat}
\affiliation{%
  \institution{Google}
  \city{Paris}
  \country{France}}
  \email{mpellat@google.com}
  
\author{Michael Strube}
\affiliation{%
  \institution{Heidelberg Institute for Theoretical Studies}
  \city{Heidelberg}
  \country{Germany}}
  \email{michael.strube@h-its.org}

\author{Lucas Dixon}
\affiliation{%
  \institution{Google}
  \city{Paris}
  \country{France}}
  \email{ldixon@google.com}

\renewcommand{\shortauthors}{L\'opez et al.}
\settopmatter{printacmref=false}
\begin{abstract}
This paper introduces a simple and effective form of data augmentation for recommender systems. A paraphrase similarity model is applied to widely available textual data -- such as reviews and product descriptions -- yielding new semantic relations that are added to the user-item graph. This increases the density of the graph without needing further labeled data.
%
The data augmentation is evaluated on a variety of recommendation algorithms, using Euclidean, hyperbolic, and complex spaces, and over three categories of Amazon product reviews with differing characteristics. 
Results show that the data augmentation technique provides significant improvements to all types of models, with the most pronounced gains for knowledge graph-based recommenders, particularly in cold-start settings, leading to state-of-the-art performance.\footnote{Code available at: \url{https://github.com/PAIR-code/recommendation-rudders}}


\end{abstract}

\ccsdesc[500]{Information systems~Recommender systems}
\keywords{recommender systems, data augmentation, knowledge graphs}

\maketitle

\section{Introduction}
\thispagestyle{empty}


Recommender systems (RS) estimate users’ preferences for items to provide personalized recommendations and a better user experience. 
The underlying assumption is that users may be interested in items selected by people who share similar interactions with them. By mining interaction records, a model \textit{implicitly} learns user-user and item-item similarities, and exploits them for recommendations \cite{salakhutdinov2008bpmf}.

One of the major challenges for RS is the long tail of users with sparse interaction data. Making recommendations for new users or about new items with little data is difficult (the so called {\em cold-start problem}). To alleviate this issue,
a considerable strand of research has explored incorporating side information to augment the interaction data \cite{zhusun2019recoswithsideinfo}.
A prominent branch of this work explores using textual descriptions and reviews, which often exist alongside rating or purchase data \cite{mcauley2013hft, tan2016ratings}.
However, a recent re-evaluation of such techniques indicates that the benefits are marginal, especially in cold-start scenarios. For example, modern deep learning-based models yield minimal changes in performance when reviews are masked \cite{sachdeva2020usefulReview}.

In this paper we introduce a simple and effective form of data augmentation that leverages pre-trained textual semantic similarity models. The models are applied to widely available textual data, like product descriptions and reviews, yielding new relations between items.
In this manner, we complement the \textit{implicit} item similarity learnt from interactions by introducing \textit{explicit} semantic relations based on textual attributes. 
We explore how these relations guide models to group semantically similar items and boost the recommendation system's performance.

The data augmentation technique is evaluated on a variety of models where the user-item graph can naturally be extended with new relations. Many of these models are variants of knowledge graph recommenders since they provide an expressive and unified framework for modelling side information between users, items, and related entities~\cite{zhang2016collkbe}. Analysis of local and global geometric measures of the generated graphs indicate that the augmented graphs are better represented in hyperbolic spaces (see \S\ref{sec:graph-analysis}). Therefore, we also explore a variety of representational alternatives for recommender systems, including Euclidean \cite{bordes2013transe, wang2014transh, yang2015distmul}, complex \cite{sun2018rotate} and hyperbolic \cite{chami2020lowdimkge, balazevic2019murp} geometries.

Finally, we analyze how the proposed relations are more efficient at encoding semantic information present in textual descriptions compared to baselines that extract latent features from raw text.
We find that our technique is more effective at reducing the generalization error, and this is particularly notable in cold-start settings.
Furthermore, we analyze which type of text is more helpful for this task, noting that the product description can provide very condensed and useful information to draw semantic relations.

In this work we investigate two types of inductive biases: a data-dependent bias, by augmenting the graph relations via pre-trained language models, and a geometric bias, through the choice of a metric space to embed the relations. By means of a thorough assessment, we show how they complement each other. As we enlarge the data, the hyperbolic properties of the graph become more evident.
In summary, we make the following contributions:
\begin{itemize}
    \item We propose an unsupervised data augmentation technique by explicitly mining semantic relations derived from items' text that boosts the performance, and is particularly effective in cold-start settings.
    \item We provide a thorough analysis of local and global structural aspects of the user-item graph, and its augmented version, which indicates that the graph is better represented in hyperbolic space rather than in Euclidean.
    \item We explore KG methods developed in Euclidean, hyperbolic and complex spaces, and showcase how they achieve state-of-the-art performance for recommendations, when leveraged with the appropriate relations.
\end{itemize}




\section{Mining Semantic Relations}
\label{sec:mining-semantic-relations}

\subsection{Motivation}
A popular approach to incorporate side information in recommender systems is to exploit user reviews and item descriptions, which often exist alongside rating or purchase data \cite{mcauley2013hft, tan2016ratings}. A textual review is much more expressive than a single rating, and the underlying assumption is that reviews are effective user/item descriptors \cite{ge2019helpfulnessaware, margaris2020reviewreliable}, thus they can be used to learn better latent features \cite{rose2017transnets, chen2018narre}.

In a conventional deep learning architecture, these features are used in matrix factorization \cite{zheng2017deepconn}.
In such a setup, the model is burdened with the task of learning an \textit{implicit} similarity function that should emerge from the textual input and the user's purchase history. Based on this affinity, 
alike users and items are grouped and leveraged for recommendations.
However, Sachdeva \& McAuley \cite{sachdeva2020usefulReview} show that recent models yield minimal changes in performance when reviews are masked. They observe that reviews are more effective when used as a regularizer \cite{mcauley2013hft, hsieh2017collaborativeML}, rather than as side data to extract latent features, and this behavior is accentuated in cold-start scenarios.

This paper introduces a different approach to benefit from textual attributes: it proposes leveraging advances in textual similarity models~\cite{cer2018use} to feed a recommender \textit{explicit} content-based similarities via new edges in the interaction graph. This requires no supervision and 
and complements the \textit{implicit} similarity function that the model learns, without increasing the computational complexity.
The data augmentation increases the density in the graph and serves as an efficient regularizer without the need to reduce the effective capacity of the model \cite{hernandez2018dataaugmentation}.


\subsection{Method}
\label{sec:semantic-relations-method}

The goal of the proposed method is to augment the user-item interaction graph with item-item relations. These relations are based on the semantic similarity between the item descriptions and reviews. 
Initially we collect all the available text for each item $i$ in the set of items $\mathcal{I}$. This includes metadata, such as item name, descriptions and reviews. We experiment with various filters, as heuristic measures of the helpfulness of different types of text (\textit{e.g.} top-k longest reviews, or only the metadata). 

To compute text embeddings we employ the Universal Sentence Encoder (USE) \cite{cer2018use}, as it has shown good performance on sentence similarity benchmarks, and it can be applied without any further fine-tuning. Moreover, the average review length in purchase datasets tends to be one paragraph \cite{ai2018explreco, mcauley2013hft}, and USE has also been pre-trained to encode paragraphs composed of more than one sentence. 
We compute one embedding $e_{i_{j}} \in \Eu^d$ for each review (or descriptor) $j$ corresponding to the item $i$. The final embedding for item $i$ is the result of taking the mean of all its review embeddings.\footnote{An ablation of the heuristics and the encoder can be found in \S\ref{sec:chunk-of-review-analysis}.}

Once we have assigned an embedding to all items, we employ \textit{cosine similarity} to compute the similarity between them. Finally, we extend the original \textit{user-item} training set with the semantic relations between pairs of items. 
We filter out low-similarity pairs, and select the top-k highest similarities to be added as relations. 



\section{Representing the Graph}



Given that we extend the user-item graph by adding semantic relations, we need to account for the different types of edges present in the augmented graph. We model this multi-relational graph as a knowledge graph, since they offer a flexible approach to add multiple relations between diverse entities. In this section we describe knowledge graphs, and their application in recommender systems.


\subsection{Knowledge Graphs}
Knowledge graphs (KGs) are multi-relational graphs where nodes represent entities and typed-edges represent relationships among entities. They are popular data structures for representing heterogeneous knowledge in the shape of \textit{(head, relation, tail)} triples, which can be queried and used in downstream applications. 
The usual approach to work with KGs is to learn representations of entities and relations as vectors, for some choice of space $\mathcal{S}$ (typically $\Eu$), such that the KG structure is preserved.
More formally, let $\mathcal{G} = (\EntitySub,\RelSub,\TripSub)$ be a knowledge graph where $\EntitySub$ is the set of entities, $\RelSub$ is the set of relations and $\TripSub \subset \EntitySub \times \RelSub \times \EntitySub$ is the set of triples stored in the graph. Most of the KG embedding methods learn vectors $\operatorname{\textbf{h}}, \operatorname{\textbf{t}} \in \Eu^{n_{\EntitySub}}$ for $h, t \in \EntitySub$, and $\textbf{r} \in \Eu^{n_{\RelSub}}$ for $r \in \RelSub$. 
The likelihood of a triple to be correct is evaluated using a model specific score function $\phi : \EntitySub \times \RelSub \times \EntitySub \rightarrow \Eu$.

\subsection{KG for Recommender Systems}
Knowledge graph embedding methods\footnote{In this work we focus on embedding methods, according to the classification presented on \cite{guo2020kgsurvey}, and not in path-based methods.} have been widely adopted into the recommendation problem as an effective tool to model side information \cite{guo2020kgsurvey}.
Multiple relations between users, items, and heterogeneous entities can be mapped into the KG and incorporated to alleviate data sparsity and enhance the recommendation performance \cite{zhang2016collkbe}.

Knowledge graph-based recommender systems can be seen as multi-task models, with well-established advantages. 
Learning several tasks (relations) at a time reduces the risk of over-fitting by generalizing the shared entity representations \cite{zhang2019dlrssurvey}.
The data augmentation also improves the generalization of the model and acts as an effective regularizer \cite{hernandez2018dataaugmentation}. 
Furthermore, akin relations can help the model to learn different types of entity interactions, such as similarities, that are finally exploited for recommendations \cite{ruder17mtl}.


Multi-relational knowledge graphs exhibit an intricate and varying structure as a result of the logical properties of the relationships they encode \cite{miller1995wordnet, Suchanek2007yago, lehmann2015dbpedia}. An item can be connected to different entities by symmetric, anti-symmetric, or hierarchical relations. To capture these non-trivial patterns more expressive operators become necessary.

In Table~\ref{tab:kg-scoring-function}, we show KG embedding methods, along with their operators, and in which RS work they have been applied. It can be seen that previous work integrating KG into RS has applied a rather narrow set of methods. These are the translational approaches \TransH{} \cite{wang2014transh}, \textsc{TransR} \cite{lin2015transr}, or \textsc{TransD} \cite{ji2015transd}, which are extensions of \TransE{} \cite{bordes2013transe}. 
We notice that the state of the art in the field of KG embedding methods has advanced in recent years, as more compound and expressive operators have been developed. However, recommender systems have continued to apply outdated models and have not profited from the current progress.

Recent approaches propose to embed the graph into non-Euclidean geometries such as hyperbolic spaces \cite{balazevic2019murp, kolyvakis2020hyperKG, chami2020lowdimkge}, to model embeddings over the complex numbers $\Comp$ \cite{trouillon2016complex, lacroix2018tensordecomp, sun2018rotate}, or apply quaternion algebra \cite{zhang2019quaternionKG}. 
In the next section we describe methods that combine different operators and achieve SotA performance on KG completion tasks.\footnote{For a brief review of hyperbolic geometry see \S\ref{sec:hyperbolic-geometry}.}


\begin{table}[!t]
\small
\centering
\adjustbox{max width=\linewidth}{
\begin{tabular}{lccl}
\toprule
\textbf{Model} & \multicolumn{2}{c}{\textbf{Scoring Function}} & \textbf{Applied in} \\
\midrule
\TransE & $-d_{\Eu}(\Head + \Rel, \Tail)$ & $\Head, \Rel, \Tail \in \Eu^n$ & \begin{tabular}[c]{@{}l@{}}\cite{he2017translationreco, tay2018LRML, huang2018seqReco, dadoun2019locreco} \\ \cite{ye2019bem, zhang2018learningOverKBE, piao2018study, ai2018explreco}\end{tabular} \\
\textsc{TransR} & $-d_{\Eu}(M_{r}\Head + \Rel, M_{r}\Tail)$ & $\Head, \Rel, \Tail \in \Eu^n$ & \begin{tabular}[c]{@{}l@{}}\cite{zhang2016collkbe, tang2019akupm, li2019kglearning}\\ \cite{wang2019kgat, sha2019akge}\end{tabular} \\
\TransH & $-d_{\Eu}(\Head_{\perp} + \Rel, \Tail_{\perp})$ & $\Head, \Rel, \Tail \in \Eu^n$ & \cite{cao2019unifying} \\
\textsc{TransD} & $-d_{\Eu}(M_{r,h}\Head + \Rel, M_{r,t}\Tail)^{2}$ & $\Head, \Rel, \Tail \in \Eu^n$ & \cite{wang2018dkn} \\
\DistMul & $\langle \Head, \Rel, \Tail \rangle$ & $\Head, \Rel, \Tail \in \Eu^n$ & \cite{xin2019rcf} \\
\RotatE & $-d_{\Eu}(\Head \circ \Rel, \Tail)$ & $\Head, \Rel, \Tail \in \Comp^n$ &  \\
\RotRef & $-d_{\Hy}(\operatorname{Attn}_{r}(\Head) \oplus \Rel, \Tail)^{2} + b_{h} + b_{t}$ & $\Head, \Rel, \Tail \in \Hy^n$ &  \\
\MuRP & $-d_{\Hy}(M_{r} \otimes \Head, \Tail \oplus \Rel)^{2} + b_{h} + b_{t}$ & $\Head, \Rel, \Tail \in \Hy^n$ & \\
\bottomrule
\end{tabular}
}
\caption{Description of different KG models}
\label{tab:kg-scoring-function}
\vspace{-4mm}
\end{table}

\subsection{Methods Compared}

\smallskip
\noindent
\RotatE{} \cite{sun2018rotate}: Maps entities and relations to the complex vector space $\Comp^n$ and defines each relation as a rotation in the complex plane from the source entity to the target entity. Given a triple (\textit{h}, \textit{r}, \textit{t}), it is expected that $\Tail \approx \Head \circ \Rel$, where $\circ$ denotes the Hadamard (element-wise) product.
Rotations are chosen since they can simultaneously model and infer inversion, composition, symmetric or anti-symmetric patterns.

\smallskip
\noindent
\MuRP{} \cite{balazevic2019murp}: By establishing a comparison with word analogies through hyperbolic distances \cite{tifrea2018poincareGlove}, the authors propose a scoring function based on relation-specific M\"obius multiplication on the head entity, and M\"obius addition \cite{ganea2018hyperNN} 
on the tail entity:
\begin{equation}
    \phi(h, r, t) = -d_{\Hy}(M_{r} \otimes \Head, \Tail \oplus \Rel)^{2} + b_{h} + b_{t}
\end{equation}
where $\Head, \Rel, \Tail \in \Hy^n$; $b_{h}, b_{t} \in \Eu$ are scalar biases for the head and tail entities respectively, and $d_{\Hy}$ is the hyperbolic distance (Eq~\ref{eq:hyper-dist}).

\smallskip
\noindent
\RotRef{} \cite{chami2020lowdimkge}: Extends \MuRP{} with rotations and reflections in hyperbolic space by learning relationship-specific isometries through Givens transformations.\footnote{\url{https://en.wikipedia.org/wiki/Givens_rotation}} The result of these operations is combined with an attention mechanism in the tangent space \cite{chami2019hgcnn}.

\begin{figure*}[!t]
\centering
\subfloat{\label{fig:ricci-music-norel}{\includegraphics[width=.25\textwidth,keepaspectratio]{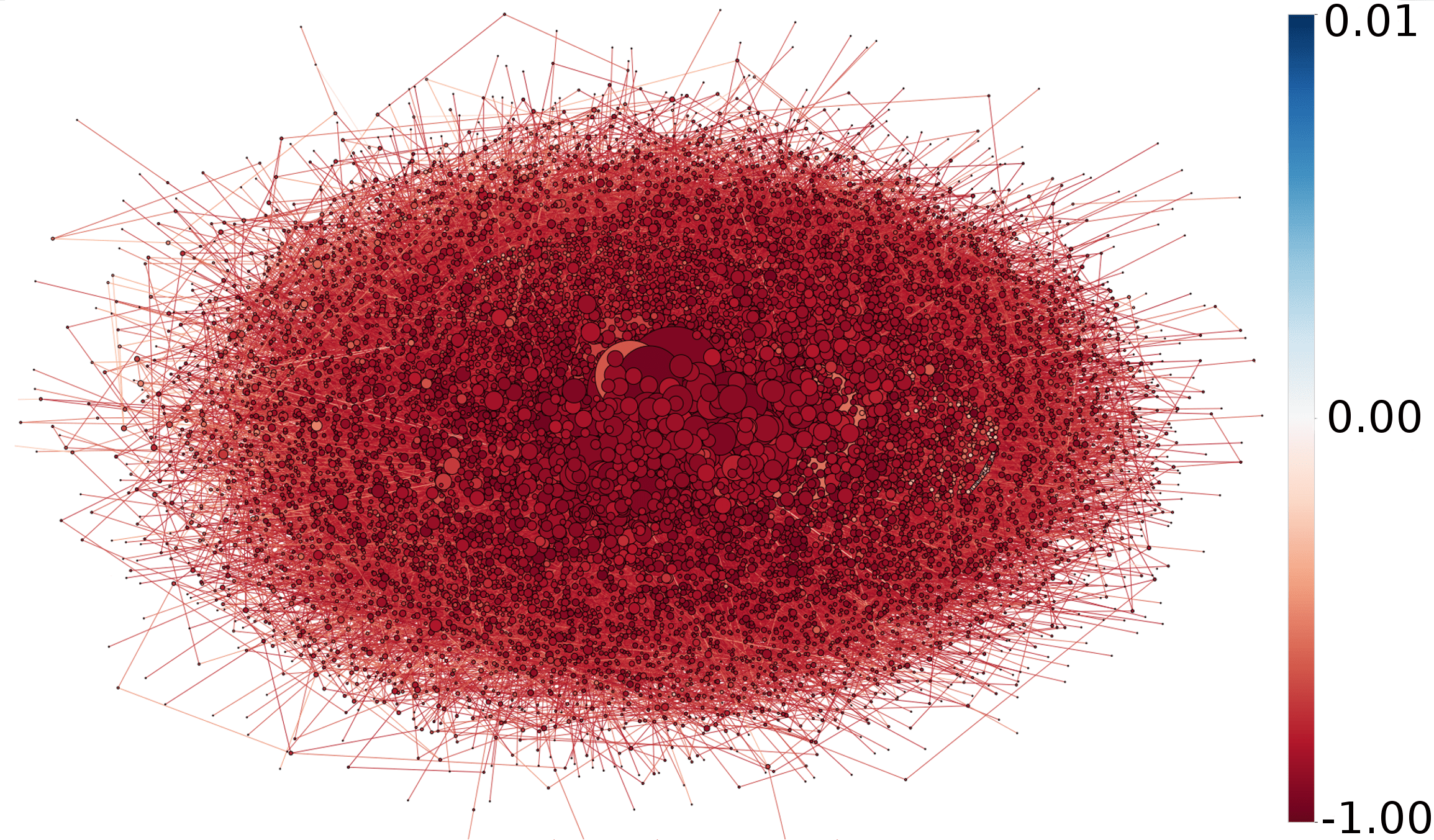}}}\hfill
\subfloat{\label{fig:ricci-music-allrel}{\includegraphics[width=.25\textwidth,keepaspectratio]{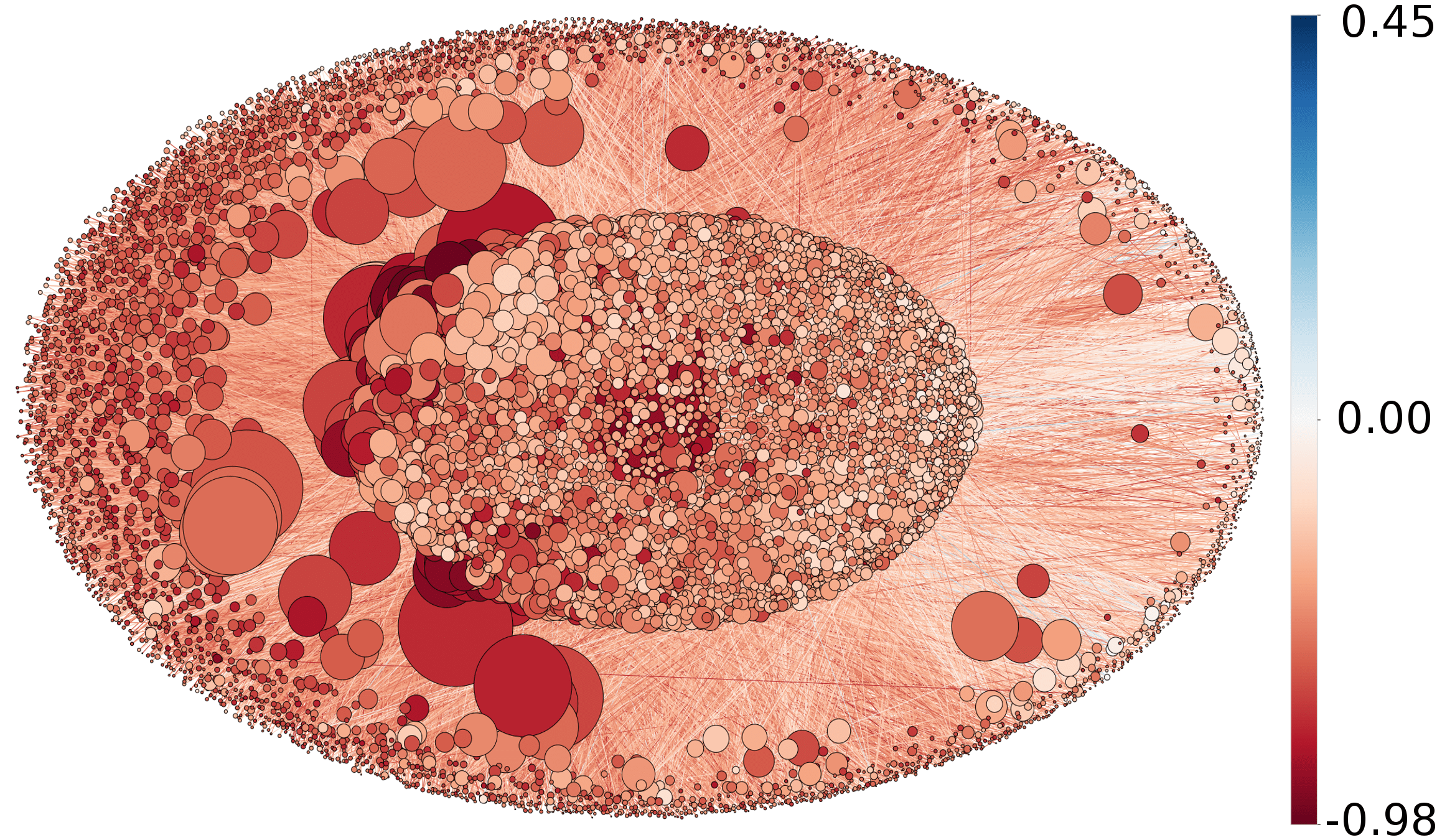}}}
\hfill
\subfloat{\label{fig:ricci-video-norel}{\includegraphics[width=.25\textwidth,keepaspectratio]{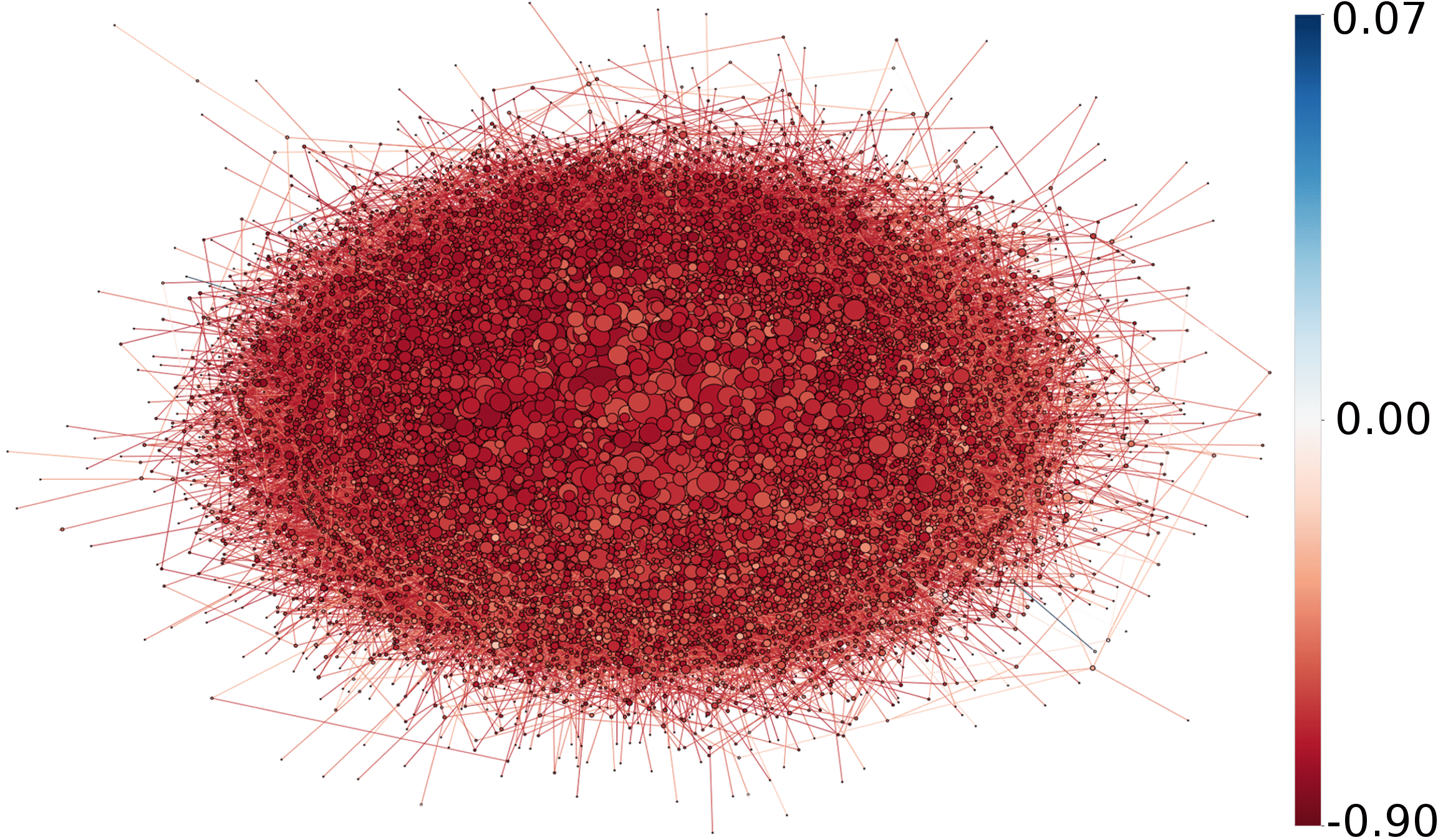}}}
\hfill
\subfloat{\label{fig:ricci-video-allrel}{\includegraphics[width=.25\textwidth,keepaspectratio]{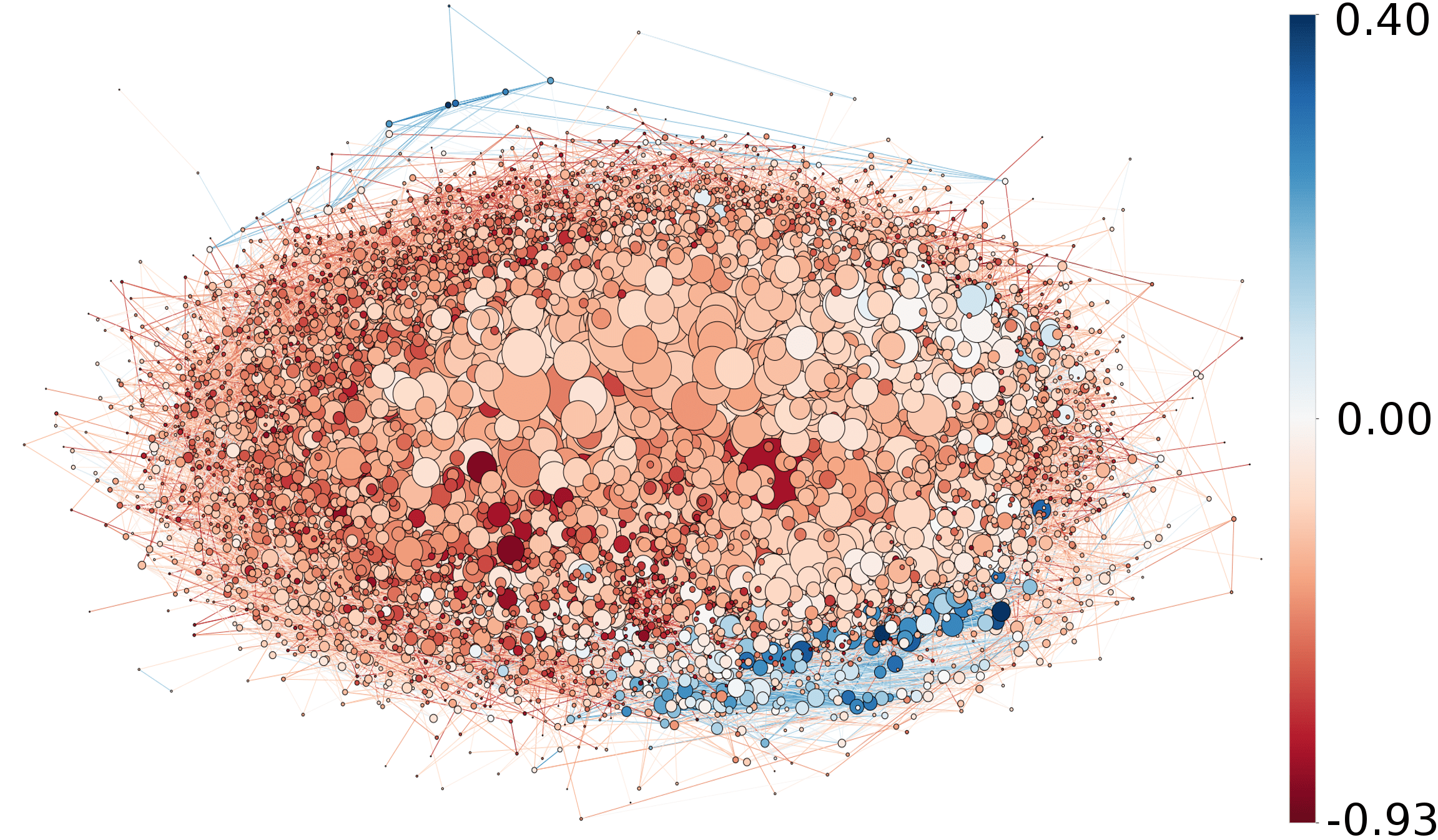}}}
\caption{Visualization of Ollivier-Ricci curvature for different graphs. Left to right: a) "MusIns" user-item, b) "MusIns" augmented graph, c) "VGames" user-item, d) "VGames" augmented graph. Node size represents node degree.
The vast majority of nodes and edges in red depict negative curvature, in correspondence with the negative curvature of hyperbolic space.
}
\label{fig:ricci-curv}
\vspace{-2mm}
\end{figure*}

\subsection{Augmenting the graph}
\label{sec:extending-the-graph}


We augment the user-item graph by exploiting heterogeneous relations between items and other entities. Our starting point is the \textit{user-item} graph, composed of solely 
$(user, buy, item)$ triples. We augment this bipartite graph with the aforementioned semantic relations. 
These are $(item\_A,$ $has\_semantic\_similarity,$ $item\_B)$ triples, meaning that there is a semantic overlap between the descriptors corresponding to these items.
Previous work has explored relations between items and diverse entities such as product brands and categories \cite{zhang2018learningOverKBE, ai2018explreco, xian2019reinforcementkg}, or movie directors and actors \cite{zhang2016collkbe, cao2019unifying, xin2019rcf}, depending on the dataset. 
To the best of our knowledge, this paper is the first to explore using pre-trained semantic similarity models to create new relations.

The graph augmentation modifies its size and structure. A change in its connectivity affects the optimal embedding space to operate \cite{gu2019lmixedCurvature}. 
Thus, we analyze this aspect in the next section.

\section{Interaction Graph Analysis}
\label{sec:graph-analysis}



The predominant approach to deal with graphs has been to embed them in an Euclidean space. Nonetheless, graphs from many diverse domains exhibit non-Euclidean features \cite{bronstein2018geomdeeplearning}. In particular, several RS datasets exhibit power-law degree distributions \cite{chamberlain2019scalableRecoSys}, and properties of scale-free networks \cite{cano2006topologyMusicRecoNets, kitsak2017latentGeomOfBipartiteNets}, which imply a latent hyperbolic geometry \cite{krioukov2010hypernetworks}. If we match the geometry of the target embedding space to the structure of the data, we can improve the representation fidelity \cite{gu2019lmixedCurvature, lopez2021sympa}.
With the goal of understanding which type of Riemannian manifold would be more suitable as embedding space, 
we analyze different structural aspects and geometric properties of the graph.

Our analysis shows that 
when we augment the relationships in the graph, the added edges modify its connectivity and structure, making it more hyperbolic-like. 



\subsection{Data}
\label{sec:data}
To investigate the recommendation problem with regard to different relationships and geometries, we focus on the Amazon dataset\footnote{\url{https://nijianmo.github.io/amazon/index.html}} \cite{mcauley2013hft, nimcauley2019extamazondata}, as it is a standard benchmark for RS, and it provides item reviews and metadata in the form of textual descriptions. Nonetheless, our analysis generalizes to RS datasets that exhibit a latent hyperbolic geometry, studied in \cite{cano2006topologyMusicRecoNets, kitsak2017latentGeomOfBipartiteNets, chamberlain2019scalableRecoSys}.

Specifically, we adopt the 5-core split for the branches "Musical Instruments" (MusIns), "Video Games" (VGames) and "Arts, Crafts and Sewing" (Arts\&Crafts), which form a diverse dataset in size and domain.
Besides the semantic relations, we also add relationships available on the dataset that have already been explored in previous work \cite{zhang2018learningOverKBE, ai2018explreco, xian2019reinforcementkg}. These are:
\begin{itemize}
    \item \textit{also\_bought}: 
    users who bought item A also bought item B. 
    \item \textit{also\_view}: 
    users who bought the item A also viewed item B.
    \item \textit{category}: 
    the item belongs to one or more categories.
    \item \textit{brand}: 
    the item belongs to one brand.
\end{itemize}
The number of each type of relation added to the final augmented graph is reported in Table~\ref{tab:data-stats}.

\begin{table}[b]
\vspace{-3mm}
\small
\centering
\adjustbox{max width=\linewidth}{
\begin{tabular}{lrrr}
\toprule
 & \multicolumn{1}{c}{\textbf{MusIns}} & \multicolumn{1}{l}{\textbf{VGames}} & \multicolumn{1}{c}{\textbf{Arts\&Crafts}} \\
\midrule
\# Users & 27,530 & 55,223 & 56,210 \\
\# Items & 10,620 & 17,408 & 22,931 \\
\hline
buy & 231,392 & 497,577 & 494,485 \\
semantic & 102,497 & 168,877 & 218,120 \\
also\_bought & 37,396 & 395,817 & 346,931 \\
also\_view & 24,811 & 278,150 & 49,512 \\
category & 45,716 & 89,273 & 85,350 \\
brand & 10,525 & 17,269 & 22,670 \\
\hline
\textbf{Total} & 452,337 & 1,446,963 & 1,217,068 \\
\bottomrule
\end{tabular}
}
\caption{Number of users, items and each type of relations added on different branches of the Amazon dataset.}
\label{tab:data-stats}
\vspace{-4mm}
\end{table}

\subsection{Hyperbolic Geometry}
\label{sec:hyperbolic-geometry}
Hyperbolic geometry is a non-Euclidean geometry with constant \textit{negative curvature}. Hyperbolic space is naturally equipped for embedding symbolic data with hierarchical structures \cite{nickel2017poincare, deSa18tradeoffs}. Intuitively, that is because the amount of space grows exponentially as points move away from the origin. This mirrors the exponential growth of the number of nodes in trees with increasing distance from the root \cite{cho2019largeMarginClassif}. Thus, hyperbolic space can be seen as the continuous analogue to a discrete tree-like structure. Embedding norm represents depth in the  hierarchy, and distance between embeddings the affinity or similarity of the respective items \cite{lopez2019figetinHS}.
In this work, we analyze models operating in the $n$-dimensional Poincar\'e ball: $\Hy^n = \{x \in \Eu^n: ||x|| < 1\}$.\footnote{Although \cite{chami2020lowdimkge} learns the curvature, we fix it to $c = 1$}.

For two points $x, y \in \Hy^n$ the distance in this space is defined as:
\begin{equation}
\small
d_{\Hy}(x, y) = \operatorname{cosh}^{-1}\left(1 + 2 \frac{\|x - y\|^2}{(1 - \|x\|^2)(1 - \|y\|^2)}\right)
\label{eq:hyper-dist}
\end{equation}

\subsection{Curvature Analysis}
Curvature is a geometric property that describes the local shape of an object. If we draw two parallel paths on a surface with positive curvature like a sphere, these two paths move closer to each other while for a negatively curved surface like a saddle, these two paths tend to be apart.
There are multiple notions of curvature in Riemannian manifolds, with varying granularity. In the interest of space\footnote{For an in-depth treatment see \cite{lee1997riemanniancurvature}.}, we only recall a key notion: hyperbolic spaces have constant negative curvature, Euclidean spaces have zero curvature (flat) and spherical spaces are positively curved.

Discrete data such as graphs do not have manifold structure. Thus, curvature analogs are necessary to provide a measure that satisfies similar properties \cite{cruceru20matrixGraph}. In this work, we apply the Ollivier-Ricci curvature to analyze the graphs \cite{ollivier2009riccicurvature}. 
Since this type of curvature characterizes the space locally, we plot the results in Figure~\ref{fig:ricci-curv}.

We can observe that nodes and edges in the user-item graphs exhibit a very negative curvature (in red color). Negatively curved edges are highly related to graph connectivity, and removing them would result in a disconnected graph \cite{ni2015riccicurvature}. As we add more relationships, the augmented graph becomes much more connected, therefore these edges play a less important role. Nonetheless, the overall curvature, as shown by the vast majority of nodes and edges, remains negative. The correspondence with the negative curvature of hyperbolic space suggests that both, the user-item and the augmented graphs would profit from a representation in that geometry, rather than in an Euclidean one.


\subsection{$\delta$-hyperbolicity}
Also known as Gromov hyperbolicity \cite{gromov1987hyperbolicity}, $\delta$-hyperbolicity quantifies with a single number the hyperbolicity of a given metric space. The smaller the $\delta$ is, the more hyperbolic-like or negatively-curved the space is. This measure has also been adapted to graphs \cite{fournier2015computingHyperbolicity}. 

We report the $\delta$-mean and $\delta$-max in Table~\ref{tab:graph-stats}. We can see that both measures decrease when we compare the user-item (U-I) graph with the augmented one (Augmen). The metric shows that, as we add more relationships to the initial user-item graph, it becomes more hyperbolic-like. This global metric complements the local curvature and also indicates that the graph fits into a hyperbolic space.

\begin{table}[b]
\small
\centering
\adjustbox{max width=\linewidth}{
\begin{tabular}{lcccccc}
\toprule
 & \multicolumn{2}{c}{\textbf{MusIns}} & \multicolumn{2}{c}{\textbf{VGames}} & \multicolumn{2}{c}{\textbf{Arts\&Crafts}} \\
 \cmidrule(lr){2-3}\cmidrule(lr){4-5}\cmidrule(lr){6-7}
 & \textbf{U-I} & \textbf{Augmen} & \textbf{U-I} & \textbf{Augmen} & \textbf{U-I} & \textbf{Augmen} \\
 \midrule
Nodes & 38,150 & 40,396 & 72,631 & 87,096 & 79,141 & 84,170 \\
Edges & 219,156 & 406,237 & 473,427 & 1,076,652 & 440,318 & 1,015,187 \\
Density (\%) & 0.015 & 0.025 & 0.009 & 0.014 & 0.007 & 0.014 \\
Avg degree & 11.5 & 20.1 & 13.0 & 24.7 & 11.1 & 24.1 \\
$\delta$-mean & 0.270 & 0.086 & 0.285 & 0.224 & 0.299 & 0.229 \\
$\delta$-max & 2 & 1.5 & 2 & 1.5 & 2 & 1.5 \\
\bottomrule
\end{tabular}
}
\caption{Statistics of graphs derived from different branches of the Amazon dataset, without relations (user-item, "U-I") and with all relations (augmented, "Augmen"). $\delta$-mean and $\delta$-max refer to the $\delta$-hyperbolicity (lower is more hyperbolic).}
\label{tab:graph-stats}
\vspace{-4mm}
\end{table}

\section{Experiments}
We experiment with different approaches to represent the multi-relational interaction graph for the task of generating recommendations. Our aim is to compare recent KG techniques, with a particular focus on the ones operating in hyperbolic spaces, with KG methods applied in previous work and SotA recommender systems.

\paragraph{Baselines:} The recommender system baselines are:
\begin{itemize}
    \item \BPR{} \cite{rendle2009bpr}: Standard collaborative filtering baseline for RS based on matrix factorization (MF) with Bayesian personalized ranking.
    \item \NeuMF{} \cite{he2017neuralCF}: Fuses MF with a multi-layer perceptron to learn latent user/item features and model interactions. We enhance this model by initializing the item features with pre-trained embeddings.
    \item \CML{} \cite{hsieh2017collaborativeML}: Metric learning baseline that encodes interactions in a joint space using Euclidean distance. Following \cite{sachdeva2020usefulReview}, we learn extra user and item biases.
    \item \HyperML{} \cite{vinh2018hyperRecommenderSystems}: Hyperbolic adaptation of \textsc{CML}.
\end{itemize}
Since reproducing all the works that adopt KGs is unfeasible (see Table~\ref{tab:kg-scoring-function}), and in cases such as \cite{zhang2016collkbe, zhang2018learningOverKBE, ai2018explreco} the KG method is applied practically without modifications, we directly employ the KG models themselves. The selected methods are:
\begin{itemize}
    \item \TransE{} \cite{bordes2013transe}: Translation-based KG method.
    \item \TransH{} \cite{wang2014transh}: Hyperplane translation-based KG method.
    \item \DistMul{} \cite{yang2015distmul}: Based on multi-linear product, a generalization of the dot product.
    \item \RotatE{} \cite{sun2018rotate}: Performs rotations in the complex plane $\Comp$.
    \item \RotRef{} \cite{chami2020lowdimkge}: Based on rotations and reflections. 
    \item \MuR{} \cite{balazevic2019murp}: Based on 
    multiplications and additions. 
\end{itemize}
For \RotRef{} and \MuR{} we compare to the Euclidean and hyperbolic versions.

\paragraph{Data}
To evaluate the models we utilize the branches of the Amazon dataset introduced in \S\ref{sec:data}. Since it is very costly to rank all the available items, following \cite{he2017neuralCF, vinh2018hyperRecommenderSystems}, we randomly select $100$ samples which the user has not interacted with, and rank the ground truth amongst these samples. To generate evaluation splits, the penultimate and last item the user has interacted with are withheld as dev and test sets respectively.

\paragraph{Implementation details:} 
Given differences in preprocessing strategies, we reproduce the baselines based on their public implementations.
To ensure consistency in the results we conduct a hyper-parameter search for all methods on the validation set. All models are trained with the Adam optimizer \cite{kingma2014Adam}, and operate with $64$ latent dimensions. We train for $1000$ epochs with early stopping on the dev set. To optimize parameters in hyperbolic models we apply tangent space optimization \cite{chami2019hgcnn}.
We choose the best learning rate for each method from from $\{0.0001, 0.0005, 0.001, 0.005\}$ and batch-size from $\{256, 512, 1024\}$. In each case, we report the average of three runs.
Preliminary experiments with multi-task losses (splitting the loss between KG and RS components as in \cite{wank2019multitaskforKG, cao2019unifying, xin2019rcf}) did not show significant improvements therefore we disregard this approach.
Models that incorporate item features (\NeuMF{} and \CML{}) are fed with the text embeddings used to compute the semantic similarities. In this way, all models have access to information extracted from the same sources.

\paragraph{Setup:} We evaluate the models in two setups: \textit{User-item}, where we only employ the user-item interactions, and \textit{Augmented}, where we utilize the graph with all the added relationships. Moreover, for the \textit{Augmented} case we follow the standard data extension protocol by adding inverse relations to the train split \cite{lacroix2018tensordecomp, chami2020lowdimkge} . That is, for each triple $(h, r, t)$, we also add the inverse $(t, r^{-1}, h)$.

\paragraph{Evaluation protocols and Metrics:}
To evaluate the recommendation performance of KG methods we only look at the \textit{buy} relation. For each user $u$ we rank the items $i_{j}$ according to the scoring function $\phi(u, buy, i_{j})$.
We adopt normalized discounted cumulative gain (DCG) 
and hit ratio (HR), both at $10$, as well-established ranking evaluation metrics for recommendations.

\paragraph{Research Questions:}
Through our experiments we aim to answer the following questions:
\begin{itemize}[label={}]
    \item \textbf{RQ1} How do KG methods perform compared to recently published RS?
    \item \textbf{RQ2} What is the impact of the data augmentation?
    \item \textbf{RQ3} How important are different relations to improving recommendations?
    \item \textbf{RQ4} Which text attributes are most helpful?
    \item \textbf{RQ5} How do different metric spaces compare?
\end{itemize}

\section{Results and Discussion}

\subsection{RQ1: Performance over user-item graph}
\label{sec:general-results}
In Table~\ref{tab:all-results} we report the results for all models.
Regarding the \textit{user-item} results, we can observe that \NeuMF{} is a very strong baseline, surpassing the performance of all other RS, and several KG methods. This can be explained by the fact that \NeuMF{} and \CML{} have access to more information than other baselines, since they are fed with embeddings generated from the items' text. 
However, \RotRef{} and \MuR{} outperform all models. 
These models are designed to deal with multi-relational graphs, but in this case they only see one type of relation ($buy$). Although they operate solely based on the user-item interactions, the compound operators that they incorporate allow them for a more expressive representation of the bipartite graph, which results in improved recommendations.
KG methods applied only on the user-item graph report a very high performance due to their enhanced representation capacity, thus we consider they should be adopted as hard-to-beat baselines in further research in recommender systems.

\begin{table*}[t]
\small
\centering
\adjustbox{max width=\textwidth}{
\begin{tabular}{lccc:ccr|cc:ccr|cc:ccr}
\toprule
 &  & \multicolumn{5}{c}{\textbf{Musical Instruments}} & \multicolumn{5}{c}{\textbf{Video Games}} & \multicolumn{5}{c}{\textbf{Arts, Crafts \& Sewing}} \\
 \cmidrule(lr){3-7}\cmidrule(lr){8-12}\cmidrule(lr){13-17} 
 &  & \multicolumn{2}{c}{\textbf{User-item}} & \multicolumn{3}{c}{\textbf{Augmented}} & \multicolumn{2}{c}{\textbf{User-item}} & \multicolumn{3}{c}{\textbf{Augmented}} & \multicolumn{2}{c}{\textbf{User-item}} & \multicolumn{3}{c}{\textbf{Augmented}} \\ \cmidrule(lr){3-4}\cmidrule(lr){5-7}\cmidrule(lr){8-9}\cmidrule(lr){10-12}\cmidrule(lr){13-14}\cmidrule(lr){15-17}
\textbf{Model} & $\mathcal{S}$ & \textbf{DCG} & \textbf{HR} & \textbf{DCG} & \textbf{HR} & $\Delta$\% & \textbf{DCG} & \textbf{HR} & \textbf{DCG} & \textbf{HR} & $\Delta$\% & \textbf{DCG} & \textbf{HR} & \textbf{DCG} & \textbf{HR} & $\Delta$\% \\
\midrule
\BPR & $\Eu$ & .266 & 43.03 & .308 & 49.00 & 13.9 & .365 & 57.67 & .397 & 59.42 & 3.0 & .320 & 48.80 & .390 & 56.79 & 16.4 \\
\HyperML & $\Hy$ & .295 & 48.72 & .264 & 47.26 & -3.0 & .329 & 53.41 & .340 & 56.31 & 5.4 & .318 & 50.57 & .322 & 53.13 & 5.1 \\
\CML & $\Eu$ & .282 & 48.82 & .345 & 55.87 & 14.4 & .323 & 55.41 & .384 & 62.31 & 12.4 & .291 & 50.47 & .358 & 59.60 & 18.1 \\
\NeuMF & $\Eu$ & .320 & 49.88 & .347 & 55.05 & 10.4 & .419 & 62.84 & .448 & 66.93 & 6.5 & .363 & 53.35 & .420 & 61.46 & 15.2 \\ \cdashline{1-1}
\TransE & $\Eu$ & .284 & 44.86 & .346 & 54.13 & 20.7 & .387 & 58.47 & .455 & 67.82 & 16.0 & .330 & 48.49 & .421 & 60.61 & 25.0 \\
\TransH & $\Eu$ & .284 & 45.05 & .342 & 54.07 & 20.0 & .390 & 58.73 & .440 & 67.51 & 14.9 & .329 & 48.22 & .422 & 61.63 & 27.8 \\
\DistMul & $\Eu$ & .251 & 40.58 & .310 & 49.75 & 22.6 & .346 & 54.85 & .370 & 57.54 & 4.9 & .290 & 44.85 & .321 & 48.61 & 8.4 \\
\RotatE & $\Comp$ & .251 & 40.22 & .332 & 53.31 & 32.6 & .345 & 54.49 & .405 & 63.80 & 17.1 & .300 & 46.37 & .389 & 58.37 & 25.9 \\
\RotRefEu & $\Eu$ & .317 & 51.63 & .366 & 57.82 & 12.0 & .408 & 63.12 & .447 & 67.31 & 6.6 & .371 & 56.43 & \textbf{.433} & 63.77 & 13.0 \\
\RotRefHy & $\Hy$ & .317 & 51.63 & \textbf{.368} & 57.86 & 12.1 & .400 & 62.78 & .450 & 68.64 & 9.3 & .363 & 55.76 & .421 & \textbf{64.09} & 14.9 \\
\MuRE & $\Eu$ & \textbf{.337} & 52.81 & .365 & \textbf{57.99} & 9.8 & .405 & 63.11 & .462 & 68.80 & 9.0 & \textbf{.372} & 56.72 & .422 & 63.95 & 12.8 \\
\MuRP & $\Hy$ & .332 & \textbf{52.84} & \textbf{.368} & 57.86 & 9.5 & \textbf{.424} & \textbf{64.92} & \textbf{.467} & \textbf{70.49} & 8.6 & .366 & \textbf{56.83} & .426 & 64.06 & 12.7 \\
\bottomrule
\end{tabular}
}
\caption{Results for "User-item" and "Augmented" graph setups, for models operating in Euclidean ($\Eu$), hyperbolic ($\Hy$) and complex ($\Comp$) space. $\Delta$\% shows the hit rate improvement over the user-item graph when data is augmented. All HR improvements are statistically significant (one-tailed Mann-Whitney U test, $p \leq 0.05$).}
\label{tab:all-results}
\vspace{-4mm}
\end{table*}

\subsection{RQ2: Exploiting Augmented Data}
\label{sec:results-augmented-data}
We can see that all models (except for \HyperML{} in MusIns) have a significant boost in performance when we train them on the \textit{augmented} graph , with gains up to $32.6\%$ for \RotatE{} over the \textit{user-item} data in MusIns. 
The proposed densification process reduces the sparsity by adding meaningful relations between the entities. 
All models, including the RS that are not designed to incorporate multi-relational information, benefit from the augmented data.
In this setup, the RS can be thought as models that aim to predict the plausibility of an interaction between any two entities (not only user-item). Although they do not account for each particular type of relation, they profit from the extended training set and this results in an enhanced generalization, which contributes to cluster users and items in a way that improves the recommendation. 


In this setting, \TransE{} and \TransH{} show a greater relative increase in their performance (higher $\Delta\%$) compared to recommender system baselines. Since the purpose of the KG models is to cope with multi-relational data, they can exploit the augmented relations in a much better way than the RS baselines. Although the rotations in the complex plane of \RotatE{} offer noticeable improvements of more than $17\%$ in all branches with the augmented graph, it is outperformed by translational approaches such as \TransE.

\HyperML{} and \CML{} do not show large gains or high performances with the added relations. The reason for this is that metric learning approaches that lack relational operators are ill-posed algebraic systems when there is a large number of interactions (see \cite{tay2018LRML}, Thm 2.1).

\MuR{} and \RotRef{} are the best performing models also in this setup.
These results highlight how heterogeneous sources of information, when leveraged through adequate tools and formulations, allow models to exploit the augmented data and achieve considerable performance gains.

Hyperbolic and Euclidean models show very competitive results for MusIns and Arts\&Crafts.
It has been shown that in low-dimensional setups ($d \in [2, 32]$) hyperbolic space offers significant improvements \cite{nickel2017poincare, leimeister2018skipGramHyper, chami2020lowdimkge}. Since we operate with $64$ dimensions, both models exhibit a similar representation capacity 
on these datasets.
Nonetheless, \MuRP{} outperforms its Euclidean counterpart in VGames for both setups.
These findings are in line with the previous analysis (\S\ref{sec:graph-analysis}), and they emphasize the importance of choosing a suitable metric space that fits the data distribution as a powerful and efficient inductive bias. Since the amount of space in the hyperbolic representation grows exponentially compared to the Euclidean one, this model is able to accommodate entities in a better way, unfolding latent hierarchies in the graph (see \S\ref{sec:metric-space-analysis}).


Finally, we also demonstrate the recent developments in KGs, and their enriched representation capacities. Advanced KG methods 
achieve a much better performance than their predecessors 
and outperform RS explicitly designed for the task in both setups. 

\begin{table*}[t]
\small
\centering
\adjustbox{max width=\textwidth}{
\begin{tabular}{lccr:ccr|ccr:ccr|ccr:ccr}
\toprule
  & \multicolumn{6}{c}{\textbf{Musical Instruments}} & \multicolumn{6}{c}{\textbf{Video Games}} & \multicolumn{6}{c}{\textbf{Arts, Crafts \& Sewing}} \\ \cmidrule(lr){2-7}\cmidrule(lr){8-13}\cmidrule(lr){14-19} 
 & \multicolumn{3}{c}{\textbf{Test}} & \multicolumn{3}{c}{\textbf{Cold Test}} & \multicolumn{3}{c}{\textbf{Test}} & \multicolumn{3}{c}{\textbf{Cold Test}} & \multicolumn{3}{c}{\textbf{Test}} & \multicolumn{3}{c}{\textbf{Cold Test}} \\ \cmidrule(lr){2-4}\cmidrule(lr){5-7}\cmidrule(lr){8-10}\cmidrule(lr){11-13}\cmidrule(lr){14-16}\cmidrule(lr){17-19}
\multicolumn{1}{c}{\textbf{Setup}} & \textbf{DCG} & \textbf{HR} & \textbf{$\Delta$\%} & \textbf{DCG} & \textbf{HR} & \textbf{$\Delta$\%} & \textbf{DCG} & \textbf{HR} & \textbf{$\Delta$\%} & \textbf{DCG} & \textbf{HR} & \textbf{$\Delta$\%} & \textbf{DCG} & \textbf{HR} & \textbf{$\Delta$\%} & \textbf{DCG} & \textbf{HR} & \textbf{$\Delta$\%} \\
 \midrule
\textit{buy} & .332 & 52.84 & - & .379 & 56.45 & - & .424 & 64.92 & - & .372 & 57.69 & - & .366 & 56.83 & - & .524 & 67.12 & - \\
\textit{buy + brand} & .351 & 54.21 & 2.6 & .418 & 58.87 & 4.3 & .425 & 64.99 & 0.1 & .404 & 61.72 & 7.0 & .371 & 58.16 & 2.3 & .533 & 68.19 & 1.6 \\
\textit{buy + category} & .348 & 54.81 & 3.7 & .376 & 61.90 & 9.6 & .426 & 65.50 & 0.9 & .399 & 61.22 & 6.1 & .370 & 57.74 & 1.6 & .546 & 68.85 & 2.6 \\
\textit{buy + also\_view} & .352 & 55.06 & 4.2 & \textbf{.449} & 62.50 & 10.7 & \textbf{.456} & 68.88 & \textbf{6.1} & .416 & 62.13 & 7.7 & .385 & 59.62 & 4.9 & .545 & 69.50 & 3.5 \\
\textit{buy + also\_bought} & .360 & 56.06 & 6.1 & .421 & 59.27 & 5.0 & .455 & \textbf{68.90} & \textbf{6.1} & .412 & 62.50 & 8.3 & \textbf{.421} & \textbf{62.92} & \textbf{10.7} & .545 & 68.49 & 2.0 \\
\textit{buy + semantic} & \textbf{.362} & \textbf{56.74} & \textbf{7.4} & .447 & \textbf{65.12} & \textbf{15.4} & .452 & 68.29 & 5.2 & \textbf{.434} & \textbf{64.97} & \textbf{12.6} & .397 & 61.01 & 7.4 & \textbf{.557} & \textbf{71.94} & \textbf{7.2} \\
\midrule
All relations & .355 & 56.81 & 7.5 & .460 & 67.44 & 19.5 & .467 & 70.49 & 8.6 & .416 & 62.73 & 8.7 & .426 & 64.06 & 12.7 & .541 & 69.68 & 3.8 \\
\bottomrule
\end{tabular}
}
\caption{Relation ablation for MuRP model, for different branches of the Amazon dataset. Results in bold show the best performing relation combination. "Cold Test" refers to the 2\% of users with the smallest number of interactions. $\Delta$\% shows the improvements of each relation with respect to only using the relation \textit{buy}.}
\label{tab:rel-ablation}
\vspace{-4mm}
\end{table*}

\subsection{RQ3: Relation Ablation}
\label{sec:results-relation-ablation}

In this section we investigate the contribution of each individual relation to the results of \MuRP{}, which we consider the best performing model. Besides evaluating on the test split, we create a subset with the $2\%$ of users with fewer number of interactions. They represent users affected by the cold-start problem, since they exhibit very few interactions with items. We refer to this split as "Cold Test". The results of the ablation are presented in Table~\ref{tab:rel-ablation}.

When we look at individual performance on the test set, we see that each relation brings improvements over the user-item graph (\textit{buy} relation), which highlights the key role of data augmentation to boost the performance of these models. 
We notice that the semantic relation is the best in the "Musical Instruments" branch, whereas in the other two branches, it is the second. The relation \textit{also\_bought} seems to be more helpful in those cases. 
\textit{also\_bought} is mined from behavioral patterns of users with respect to \textit{complementary} products that are usually bought together \cite{mcauley2015complementarySuplementary}, and it plays a fundamental role in predicting user purchases \cite{wolbitsch2019beggars}. However, the relations \textit{also\_bought} and \textit{also\_view} are derived from the entire dataset, and not only from the train split. These relations incorporate information between users and items on the dev/test set, posing considerable advantages over the \textit{brand}, \textit{category} and \textit{semantic} relations.\footnote{This was confirmed by personal correspondence with the authors of \cite{nimcauley2019extamazondata}.}

Finally, all relationships combined outperform the individual setups for all branches of the dataset in the test set, in line with previous results \cite{zhang2018learningOverKBE}. This demonstrate how KG methods can leverage heterogeneous information in an unified manner, and shows the scalability of the approach to new relation types.



\paragraph{Cold-start Problem}

We analyze how different relations affect users in cold-start settings, by looking at the $2\%$ of users with the fewest number of interactions (Cold Test).
In this case, the \textit{semantic} relation brings a very pronounced boost in performance: $15.4\%$, $12.6\%$ and $7.2\%$ for MusIns, VGames and Arts\&Crafts respectively, much more than any other relation.
This shows the effectiveness of semantic relations to densify the graph, with remarkable improvements particularly over sparse users and items. 
These observations are in line with previous research that has shown how reviews, when used as regularizers, are particularly helpful to alleviate cold-start problems \cite{sachdeva2020usefulReview}. 
Moreover, we notice that the performance for "Cold Test" tends to be better than for "Test" in MusIns and Arts\&Crafts. We hypothesize that this is caused by modelling users as a single point in the space. When users exhibit a large number of interactions with distinct items, it becomes more difficult to place the user embedding close to all their preferences. 


\subsection{RQ4: The Role of Reviews}
\label{sec:chunk-of-review-analysis}



\subsubsection{Relations vs Features}
We analyze the effectiveness of semantic relations to model side information extracted from textual descriptions when compared to different ways of incorporating latent features proposed in previous work.
We consider the following models:
\CML{} utilizes the item features as an explicit regularizer, \NeuMF{} initializes the item embeddings with the features, and \Narre{} \cite{chen2018narre} uses TextCNN \cite{kim2014textCNN} to extract features from the text. We do not compare to KG approaches since they do not incorporate latent features. For \CML{} and \NeuMF{} we employ the text embeddings used to compute the semantic similarities as item features. 
Results for the ablation are reported in Table~\ref{tab:adding-info}.

Compared to baselines that do not incorporate any side information, we see that relations bring a larger improvement than features for \CML{} and \Narre{} (17.1 vs 3.7 and 10.6 vs 7.4 relative improvement respectively).  Although these models are not specifically designed to incorporate multi-relational information, data augmentation via semantic relations seems to be more effective than exploiting textual features.
For \NeuMF{}, the transfer learning technique of initializing the item embeddings with textual features proves to be very effective. 
Nonetheless, when semantic relations are combined with features, the performance is improved.

This ablation showcases the efficacy of semantic relations to model information extracted from textual data. They can be seamlessly integrated with latent features, and the improvements of the combined models demonstrate that they provide complementary information for recommendations.

\begin{table}[b]
\vspace{-3mm}
\small
\centering
\adjustbox{max width=\linewidth}{
\begin{tabular}{lcccccc}
\toprule
\textbf{MusIns} & \multicolumn{2}{c}{\CML} & \multicolumn{2}{c}{\NeuMF} & \multicolumn{2}{c}{\Narre} \\
\cmidrule(lr){2-3}\cmidrule(lr){4-5}\cmidrule(lr){6-7}
\textbf{Setup} & \textbf{HR} & \textbf{$\Delta$} & \textbf{HR} & \textbf{$\Delta$} & \textbf{HR} & \textbf{$\Delta$} \\
\midrule
\xmark~\hspace{0.7mm}feat, \xmark~\hspace{0.7mm}rel & 47.46 & - & 41.50 & - & 46.09 & - \\
\xmark~\hspace{0.7mm}feat, \checkmark~rel & 55.57 & 17.1 & 44.13 & 6.3 & 50.98 & 10.6 \\
\checkmark~feat, \xmark~\hspace{0.7mm}rel & 49.22 & 3.7 & 49.64 & 19.6 & 49.50 & 7.4 \\
\checkmark~feat, \checkmark~rel & 53.95 & 13.7 & 50.80 & 22.4 & 51.22 & 11.1 \\
\bottomrule
\end{tabular}
}
\caption{Ablation of adding item features and/or semantic relations for the MusIns branch.}
\label{tab:adding-info}
\vspace{-4mm}
\end{table}

\subsubsection{Type of Text for Semantic Relation}
We analyze which type of text is more useful for extracting features and capturing item similarities. To do so, we filter the available text for each item according to different criteria: (a) Only textual metadata, such as product name, description and categorical labels, (b) only reviews, (c) only reviews with high level of sentiment polarity, and (d) metadata + top-k longest reviews.
We repeat the method described in \S\ref{sec:semantic-relations-method} to pre-process relations, and encode all text with USE. We compare \CML{} and \NeuMF, which are models that incorporate item features, and \MuRP. In all cases we run experiments with the \textit{buy + semantic} training set, in order to analyze the addition of semantic information only. Results for "MusIns" are reported in Table~\ref{tab:chunk-of-review}.
We can see that adding features on top of relations degrade the performance of \CML{} (the same behavior can be noticed in Table~\ref{tab:adding-info}). This model integrates item features as a regularizer to correct the projection into the target embedding space.
In this setup, the semantic relations are already contributing in that regard, therefore the integration of extra features seems to mislead the model. On the other hand, \NeuMF{} has a drastic drop in performance when features are removed. In this case, the features and relations extracted from text that combine metadata with the longest reviews helps the most in the recommendations, followed by utilizing all the available reviews as input.

Finally, \MuRP{} does not use any features and solely learns from the relations available in the graph. We again see that using metadata combined with the longest reviews is the most useful text to learn item dis/similarities. This is due to the fact that longer reviews tend to be more descriptive about the items. However, results using only metadata show competitive performance as well. In the three models, reviews with high polarity (i.e. very "positive/negative" reviews rather than "neutral" ones) do not seem to carry extra descriptive information such that it can be leveraged for similarities.

This ablation suggests that to learn useful dis/similarities between items for recommendations, it is convenient to combine the longest reviews with the item metadata.
Nonetheless, it is noticeable that metadata by itself 
can be leveraged to obtain competitive results. We consider this a relevant outcome since, in the absence of lengthy reviews, a brief and accurate item name and description might as well offer remarkable performance gains.

\begin{table}[t]
\small
\centering
\adjustbox{max width=\linewidth}{
\begin{tabular}{lcccccc}
\toprule
\textbf{MusIns} & \multicolumn{2}{c}{\CML} & \multicolumn{2}{c}{\NeuMF} & \multicolumn{2}{c}{\MuRP} \\
 \cmidrule(lr){2-3}\cmidrule(lr){4-5}\cmidrule(lr){6-7}
\textbf{Setup} & \textbf{DCG} & \textbf{HR} & \textbf{DCG} & \textbf{HR} & \textbf{DCG} & \textbf{HR} \\
\midrule
No features & \textbf{.346} & \textbf{55.57} & .273 & 44.13 & - & - \\
Metadata & .287 & 53.17 & .310 & 49.33 & .349 & 55.29 \\
Reviews & .334 & 53.95 & .316 & 50.80 & .354 & 55.55 \\
Polarity revs. & .321 & 51.68 & .305 & 49.17 & .344 & 54.54 \\
Meta + LongRevs & .337 & 54.80 & \textbf{.320} & \textbf{51.06} & \textbf{.362} & \textbf{56.74} \\
\bottomrule
\end{tabular}
}
\caption{Comparison of models that exploit textual information extracted from different types of text. All models are trained over the user-item graph + semantic relations.}
\label{tab:chunk-of-review}
\vspace{-4mm}
\end{table}

\subsubsection{Encoder Analysis}
This ablation is related to the choice of a pre-trained encoder that captures text similarities in an unsupervised manner. We compare the performance of USE to BERT \cite{devlin-etal-2019-bert}, without applying any fine-tuning, and Sentence-BERT \cite{reimers2019sentencebert}, which is an adaptation of BERT with a siamese network, fine-tuned for sentence similarity. We analyze the \MuRP{} model with the same four criteria for filtering the available text, and we compare to the setup without adding the semantic relations as well.

\begin{figure}[b]
\vspace{-3mm}
\centering
\includegraphics[clip, trim=4.6cm 0.8cm 3.2cm 1.2cm, width=\linewidth, height=2.9cm]{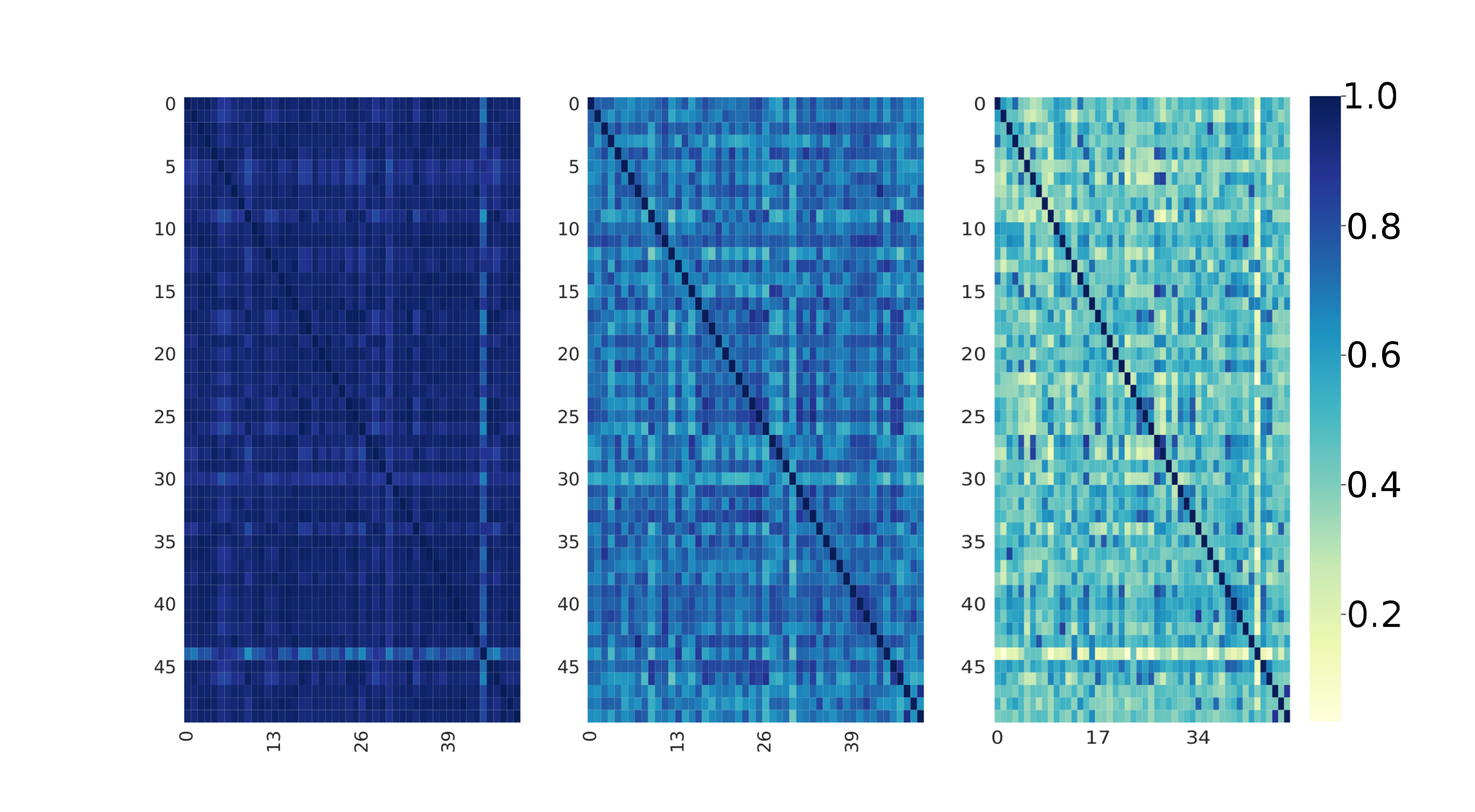}
\caption{Cosine similarity of BERT (left), S-BERT (center) and USE (right) review embeddings for 50 items.}
\label{fig:sym-rels}
\vspace{-4mm}
\end{figure}

Results for the "MusIns" are reported in Table~\ref{tab:encoder-ablation}. We observe that when we use BERT as the encoder, the performance of the user-item graph extended with semantic relations is worse than using the user-item graph alone. In Figure~\ref{fig:sym-rels} we show the cosine similarities for $50$ random item embeddings. We see that for the BERT encoder, which has not been pre-trained on semantic similarity objectives, most items are very similar to each other, and this hinders the model from clustering them. This setup can be considered as an ablative experiment of the original model, where we add random semantic relations. This demonstrates the importance of the semantic information contained in the connections that we create, and how the model is able to leverage them. 

Sentence-BERT shows improvements over BERT. Nevertheless, we find USE to be the most effective encoder that captures review and metadata dis/similarities. Sentence-BERT is more competent at distinguishing degrees of similarity than BERT, but Figure~\ref{fig:sym-rels} shows that the patterns detected are alike. On the other hand, USE is able to identify a broader range of dis/similarities between items, that results in an improved performance for recommendations. 


\begin{table}[t]
\small
\centering
\adjustbox{max width=\textwidth}{
\begin{tabular}{lcccccc}
\toprule
\textbf{MusIns} & \multicolumn{2}{c}{\textbf{BERT}} & \multicolumn{2}{c}{\textbf{S-BERT}} & \multicolumn{2}{c}{\textbf{USE}} \\
 \cmidrule(lr){2-3}\cmidrule(lr){4-5}\cmidrule(lr){6-7}
\textbf{Setup} & \textbf{DCG} & \textbf{HR} & \textbf{DCG} & \textbf{HR} & \textbf{DCG} & \textbf{HR} \\
\midrule
User-item & \textbf{.332} & \textbf{52.84} & .332 & 52.84 & .332 & 52.84 \\
Metadata & .331 & 52.80 & .334 & 53.43 & .349 & 55.29 \\
Reviews & .325 & 51.99 & \textbf{.341} & 54.18 & .354 & 55.55 \\
Polarity revs. & .310 & 50.46 & .331 & 52.46 & .344 & 54.54 \\
Meta + LongRevs & .328 & 52.35 & .339 & \textbf{54.31} & \textbf{.362} & \textbf{56.74} \\

\bottomrule
\end{tabular}
}
\caption{Performance of MuRP model on MusIns for different encoders, and different filtering heuristics.}
\label{tab:encoder-ablation}
\vspace{-4mm}
\end{table}

\subsection{RQ5: Metric Space Analysis}
\label{sec:metric-space-analysis}
Our experiments showed that hyperbolic methods 
can offer improvements over systems operating on Euclidean space or with complex numbers.
Moreover, since hyperbolic space is naturally equipped for embedding hierarchical structures, its self-organizing properties make it amenable to capture different types of hierarchies as a by-product of the learning process. In the resulting embeddings, the norm represents depth in the hierarchy.
As explained in \S\ref{sec:extending-the-graph}, each item in the Amazon dataset has a set of categorical labels 
that describes which categories the item belongs to. 
We analyze the hyperbolic and Euclidean versions of \MuR, and report the Spearman correlation between the norms of the embeddings for each category and the number of interactions of that category:
\begin{itemize}
    \item Hyperbolic: $-0.52$
    \item Euclidean: $-0.06$
\end{itemize}
The correlation is moderate to high for the hyperbolic model, whereas for the Euclidean model it is non-existent. This indicates that more "general" categories (with more interactions) have a shorter norm, 
which is expected when embedding a hierarchy in a hyperbolic space \cite{deSa18tradeoffs}, while the origin of the space has no particular meaning in the Euclidean model.

To shed light on this aspect, we reconstruct the hierarchies that the hyperbolic and Euclidean models learn. To do so, we randomly select a category embedding, and iteratively look at the closest neighbor that has less or equal norm (this would be the parent category). We report the hierarchies for "Electric Guitar Bags \& Cases" and "Footswitches" from the "MusIns" dataset in Table~\ref{tab:hyperbolic-categories}.

We notice that the hyperbolic hierarchy is much more concise and precise, compared to the Euclidean one. The hyperbolic model builds different "short" hierarchies, accommodating the labels in a more spread way, whereas the Euclidean model learns a "tall" tree of categories, which results in a much more noisy arrangement.
 
This shows that the hyperbolic model automatically infers the hierarchy arising from the label distribution \cite{lopez2019figetinHS, lopez2020fullyhyper}, and offers a more interpretable space. Furthermore, the model achieves this as a by-product of the learning process with augmented relations, without being specifically trained for this purpose.

\begin{table}[b]
\vspace{-4mm}
\small
\centering
\adjustbox{max width=\linewidth}{
\begin{tabular}{clcl|clcl}
\toprule
\multicolumn{2}{c}{\textbf{Hyperbolic}} & \multicolumn{2}{c}{\textbf{Euclidean}} & \multicolumn{2}{c}{\textbf{Hyperbolic}} & \multicolumn{2}{c}{\textbf{Euclidean}} \\
\cmidrule(lr){1-2}\cmidrule(lr){3-4}\cmidrule(lr){5-6}\cmidrule(lr){7-8}
\textbf{Norm} & \textbf{Label} & \textbf{Norm} & \textbf{Label} & \textbf{Norm} & \textbf{Label} & \textbf{Norm} & \textbf{Label} \\
\midrule
0.42 & \begin{tabular}[c]{@{}l@{}}Musical \\ Instruments\end{tabular} & 1.47 & \begin{tabular}[c]{@{}l@{}}Musical \\ Instruments\end{tabular} & 0.42 & \begin{tabular}[c]{@{}l@{}}Musical \\ Instruments\end{tabular} & 1.47 & \begin{tabular}[c]{@{}l@{}}Musical \\ Instruments\end{tabular} \\
\hline
0.77 & Bags \& Cases & 2.65 & DJ Sets & 0.76 & \begin{tabular}[c]{@{}l@{}}Amplifier \\ Accessories\end{tabular} & 2.57 & \begin{tabular}[c]{@{}l@{}}Orchestral Strings \\ Accessories\end{tabular} \\
\hline
0.83 & \begin{tabular}[c]{@{}l@{}}Bass Guitar \\ Bags \& Cases\end{tabular} & 2.70 & CD Players & 0.82 & \begin{tabular}[c]{@{}l@{}}Footswitches \& \\ Controllers\end{tabular} & 2.67 & Cello \\
\hline
0.85 & \begin{tabular}[c]{@{}l@{}}Electric Guitar \\ Bags \& Cases\end{tabular} & 2.96 & Bags \& Cases & 0.84 & Footswitches & 2.72 & Viola \\
\hline
\multicolumn{1}{l}{} &  & 2.99 & \begin{tabular}[c]{@{}l@{}}Roomy exterior \\ pockets\end{tabular} & \multicolumn{1}{l}{} &  & 2.99 & Parts \\
\hline
\multicolumn{1}{l}{} &  & 3.03 & 25mm padding & \multicolumn{1}{l}{} &  & 3.00 & \begin{tabular}[c]{@{}l@{}}Amplifier \\ Accessories\end{tabular} \\
\hline
\multicolumn{1}{l}{} &  & 3.15 & \begin{tabular}[c]{@{}l@{}}Bass Guitar \\ Bags \& Cases\end{tabular} & \multicolumn{1}{l}{} &  & 3.10 & \begin{tabular}[c]{@{}l@{}}Footswitches \& \\ Controllers\end{tabular} \\
\hline
\multicolumn{1}{l}{} &  & 3.33 & \begin{tabular}[c]{@{}l@{}}Electric Guitar \\ Bags \& Cases\end{tabular} & \multicolumn{1}{l}{} &  & 3.25 & Footswitches \\
\bottomrule
\end{tabular}
}
\caption{Category hierarchies and norm of each category label for hyperbolic and Euclidean spaces.}
\label{tab:hyperbolic-categories}
\vspace{-5mm}
\end{table}

\section{Related Work}
\paragraph{Data Augmentation:}
Data augmentation plays an important role in machine learning \cite{krizhevsky2012, dataaugmentationsurvey}, as it reduces the generalization error without affecting the effective capacity of the model \cite{hernandez2018dataaugmentation}. 
In RS, augmentation techniques have been applied by extending co-purchased products \cite{wolbitsch2019beggars}, generating adversarial pseudo user-item interactions \cite{wang2019adversarialAugmentation}, or casting different user actions as purchases \cite{tan2016rnnSessionAugmented, tuan2017cnnsession}. Also by exploiting item side information, such as images \cite{chu2017visualReco}, audio \cite{LiangZE15} or video \cite{chen2018videoReco} features.
We propose a new unsupervised method to learn similarity relations between items (or users) based on semantic text models applied to textual attributes.
Our work expands the strand of research that incorporates review information as regularization technique~\cite{sachdeva2020usefulReview}, and notably improves the performance for the cold-start.

\paragraph{Recommenders using Text:} Previous work has mined text to use it as regularizer~\cite{mcauley2013hft, hsieh2017collaborativeML}, or as latent features to learn better user and item representations~\cite{zheng2017deepconn, chen2018narre}. 
However, Sachdeva \& McAuley 
argue that the benefit of using reviews for recommendation is overstated, and the reported gains are only possible under a narrow set of conditions~\cite{sachdeva2020usefulReview}. They specifically note that reviews seem to provide little benefit as features, but help more when used for regularization. The simple data augmentation method proposed in this paper offers another way to think about using textual information. 
The added relations improve user and item representations, like adding features can, but without needing to increase the representation size. Thus it can also be seen as providing the benefits of regularization, but without directly constraining model expressivity, as for instance dropout does.

\paragraph{Knowledge Graph Recommenders:} Previous work integrating KG into RS has applied a narrow set of representational methods, favoring Euclidean translational approaches \cite{guo2020kgsurvey} (see Table~\ref{tab:kg-scoring-function}). The experiments in this paper expand on previous analysis by including more recent KG embedding methods. Our results show that newer methods significantly benefit from the introduced data augmentation and outperform not only previous KG recommenders, but also other state-of-the-art recommendation systems.

\paragraph{Hyperbolic Space:} The advantages of hyperbolic space have been argued for in a wide variety of application domains: question answering \cite{tay2018hyperQA}, machine translation \cite{gulcehre2018hyperAttentionNet}, language modeling  \cite{dhingra2018embeddingTextInHS}, hierarchical classification \cite{lopez2019figetinHS, lopez2020fullyhyper}, and taxonomy refinement \cite{aly2019everyChild, le2019inferringHearstPatterns} among others. 
In RS, hyperbolic geometry naturally emerges in several datasets \cite{cano2006topologyMusicRecoNets, kitsak2017latentGeomOfBipartiteNets}, and hyperbolic spaces have been applied in combination with metric learning approaches \cite{vinh2018hyperRecommenderSystems, chamberlain2019scalableRecoSys}.
In this work, we expand on these studies and carry out a structural analysis of the properties of user-item graphs extended by our data augmentation, which shows that hyperbolic methods may have an important role to play for both interpretability of recommendations, and for enabling higher performance by leveraging lower representational dimensionality.

\section{Conclusions}



In this work we propose a simple unsupervised data augmentation technique that adds semantic relations to the user-item graph based on applying pre-trained language models to widely available textual attributes. This can be regarded as a data-dependent prior that introduces an effective inductive bias, without increasing the computational cost of models at inference time.

By exploring a variety of modern KG methods, we observe that recent advances, when combined with our data augmentation technique, result in state-of-the-art RS performance (RQ1, \S\ref{sec:general-results}). 
Moreover, the proposed data augmentation improves the performance of all analyzed models, including those that are not designed to handle multi-relational information (RQ2, \S\ref{sec:results-augmented-data}). Thus, the technique can be considered architecture-agnostic.

Our ablation study highlights the impact of the semantic relations particularly in cold-start settings (RQ3, \S\ref{sec:results-relation-ablation}). Regarding the choice of textual inputs, the study reveals that using either reviews or brief product descriptions are both effective (RQ4, \S\ref{sec:chunk-of-review-analysis}).
An important branch of further work here is to explore if these results generalise to denser domains, as we notice anecdotal evidence that the benefit of this data augmentation diminishes as the average degree of nodes increases. 

Finally, our analysis of structural properties of the graphs, which can be extended to more datasets that exhibit a latent hyperbolic geometry \cite{cano2006topologyMusicRecoNets, kitsak2017latentGeomOfBipartiteNets, chamberlain2019scalableRecoSys}, reveals that recommenders can benefit from operating in these metric spaces. In particular, we remark how hyperbolic space improves the interpretability for recommendations (RQ5, \S\ref{sec:metric-space-analysis}).




\begin{acks}
This work has been supported by the German Research Foundation (DFG) as part of the Research Training Group Adaptive Preparation of Information from Heterogeneous Sources (AIPHES) under grant No. GRK 1994/1 and the Klaus Tschira Foundation, Heidelberg, Germany.
\end{acks}

\bibliographystyle{plain}
\bibliography{references}


\end{document}